\documentclass[letterpaper]{article} 
\usepackage{aaai25}  
\usepackage{times}  
\usepackage{helvet}  
\usepackage{courier}  
\usepackage[hyphens]{url}  
\usepackage{graphicx} 
\usepackage{natbib}  
\usepackage{caption} 
\usepackage{algorithm}
\usepackage{algorithmic}

\usepackage{booktabs}
\usepackage{multirow}
\usepackage{graphicx}
\usepackage{amsmath,amssymb,stmaryrd}  

\usepackage{newfloat}
\usepackage{listings}
\DeclareCaptionStyle{ruled}{labelfont=normalfont,labelsep=colon,strut=off} 
\floatstyle{ruled}
\newfloat{listing}{tb}{lst}{}
\floatname{listing}{Listing}
\title{Learning from Mistakes: Self-correct Adversarial Training for Chinese Unnatural Text Correction}
\author{
    Xuan Feng\textsuperscript{\rm 12},
    Tianlong Gu\textsuperscript{\rm 12}\thanks{Corresponding author.},
    Xiaoli Liu\textsuperscript{\rm 123},
    Liang Chang\textsuperscript{\rm 4}
}
\affiliations{
    \textsuperscript{\rm 1}College of Cyber Security, Jinan University\\
    \textsuperscript{\rm 2}Engineering Research Center of Trustworthy AI (Ministry of Education)\\
    \textsuperscript{\rm 3}Graduate School of Engineering, Chiba University\\
    \textsuperscript{\rm 4}Guangxi Key Laboratory of Trusted Software\\
    fenffef@163.com, $\{$gutianlong, txlliu$\}$@jnu.edu.cn, changl@guet.edu.cn
}

\usepackage{bibentry}

\begin{document}

\maketitle

\begin{abstract}
Unnatural text correction aims to automatically detect and correct spelling errors or adversarial perturbation errors in sentences. Existing methods typically rely on fine-tuning or adversarial training to correct errors, which have achieved significant success. However, these methods exhibit poor generalization performance due to the difference in data distribution between training data and real-world scenarios, known as the exposure bias problem. In this paper, we propose a self-correct adversarial training framework for \textbf{L}earn\textbf{I}ng from \textbf{MI}s\textbf{T}akes (\textbf{LIMIT}), which is a task- and model-independent framework to correct unnatural errors or mistakes. Specifically, we fully utilize errors generated by the model that are actively exposed during the inference phase, i.e., predictions that are inconsistent with the target. This training method not only simulates potential errors in real application scenarios, but also mitigates the exposure bias of the traditional training process. Meanwhile, we design a novel decoding intervention strategy to maintain semantic consistency. Extensive experimental results on Chinese unnatural text error correction datasets show that our proposed method can correct multiple forms of errors and outperforms the state-of-the-art text correction methods. In addition, extensive results on Chinese and English datasets validate that LIMIT can serve as a plug-and-play defense module and can extend to new models and datasets without further training.
\end{abstract}

\section{Introduction}
Unnatural text correction (UTC) is a task that automatically corrects a variety of textual errors or mistakes in a given sentence, including spelling errors (such as visual and phonetic errors), and adversarial perturbation errors. It has attracted much attention in academia and industry due to the important role of UTC in improving text accuracy and readability \cite{Liu_Wu_Zhao_2024}. With the widespread of unnatural texts and euphemisms on the Internet, it has become increasingly significant in various domains \cite{feng-2024-protect}. For example, UTC can automatically fix errors in user-generated content and improve the quality of content moderation \cite{pmlr-v202-dai23c}. Moreover, it is possible for UTC to detect and correct adversarial perturbations, enhancing the robustness and trustworthiness of the system \cite{liu2023twins}. Therefore, correcting unnatural text errors or mistakes is crucial for content moderation and robust defense.

\begin{figure}[t]
\centering
\includegraphics[width=\linewidth]{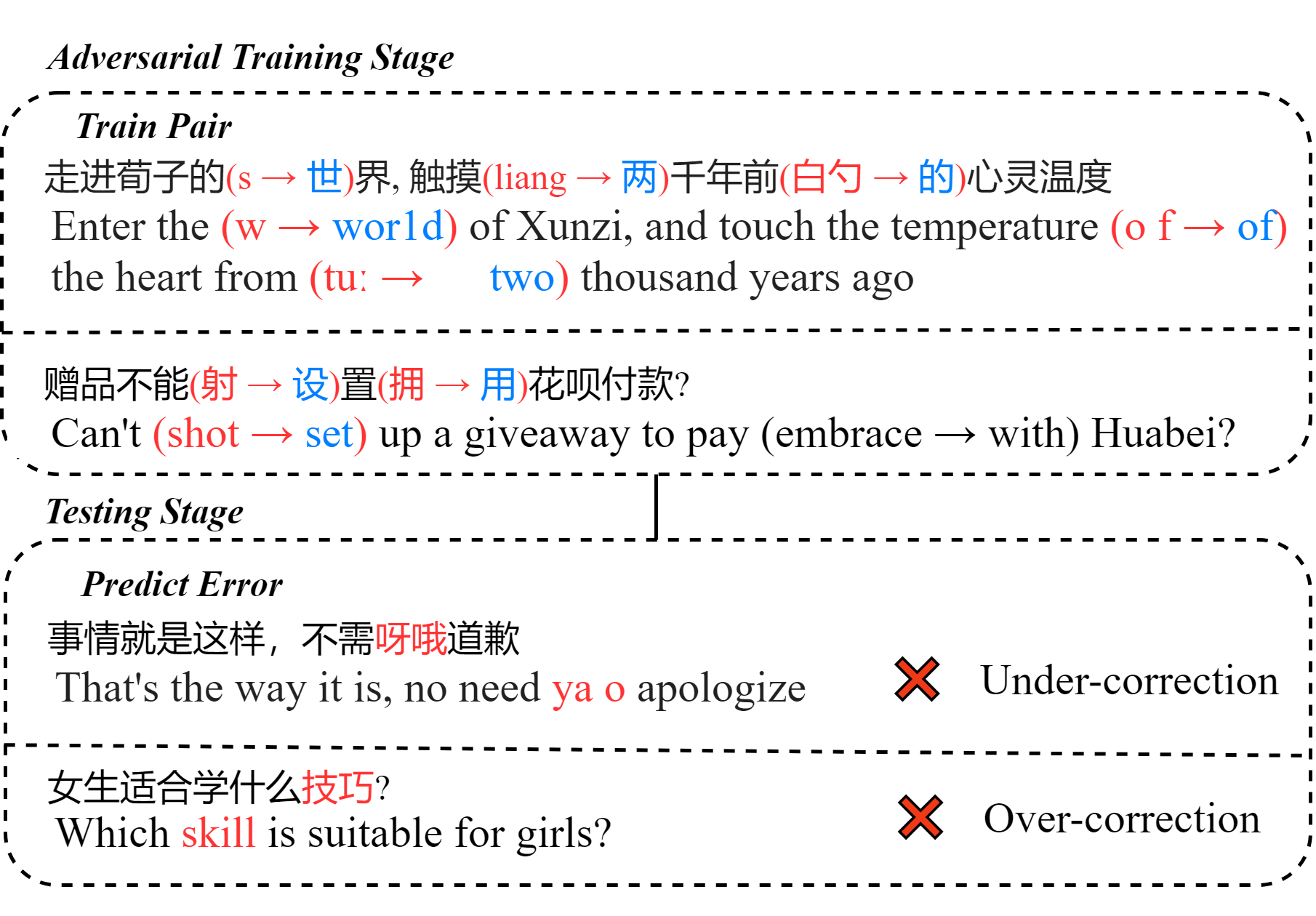}
\caption{Examples of various unnatural text error types, the red characters are characters with errors, while the blue characters are correct characters. For easier understanding, pinyin errors in Chinese are represented by phonetic symbols in English (two $\to$ tu:).}
\label{fig:1}
\end{figure}

Existing text correction methods typically rely on fine-tuning or adversarial training paradigms \citet{liu2021plome, li-etal-2022-learning-dictionary, li-etal-2022-past, li-etal-2022-improving-chinese, wu-etal-2023-rethinking, Liu_Wu_Zhao_2024}. Although these methods have achieved significant success in common text correction tasks, they often exhibit poor generalization performance in real-world applications \cite{gupta-etal-2023-dont}. As shown in Figure 1, when encountering unnatural text errors, the model may be under-corrected or over-corrected. On the one hand, the under-correction is mainly due to the difference in data distribution between the training data and the real-world scenario, which is known as the exposure bias problem \cite{bengio2015scheduled}. Specifically, training data are usually constructed or preprocessed manually with a relatively fixed and uniform distribution, while real-world data distributions are more complex and variable. In this case, the patterns and representations learned during training may not be sufficient to serve in the reasoning process, leading to unsatisfactory corrections. Thus, effectively solving the exposure bias problem and enhancing the model's generalization ability in real-world applications has become an important challenge.
On the other hand, the model will also over-correct characters without errors in the text \cite{liang-etal-2023-disentangled}. In general, over-correction will distort the original meaning of the text and affect the reader's understanding. It can also affect the user's trust and experience with the error correction system. Therefore, maintaining semantic consistency while correcting text is also an important concern.

In these regards, we propose a self-correct adversarial training framework for \textbf{L}earn\textbf{I}ng from \textbf{MI}s\textbf{T}akes (\textbf{LIMIT}), which effectively copes with multi-type spelling errors and adversarial perturbations without external knowledge effectively. Specifically, we first implement a generative correction mechanism that enables models to correct multi-type errors or mistakes. As a unified mechanism, it corrects adversarial perturbations that are specific to different models and tasks. Second, we introduce self-correct adversarial training to fine-grain the contrasting examples according to the ranking loss, thereby obtaining robust representations. During the training process, incorrect examples generated based on the model's own predictions (e.g., samples inconsistent with the target generated by a beam search algorithm) are also incorporated into the learning process. This training process motivates the model to identify and correct its own biases by actively exposing its prediction errors in the inference phase. It not only mitigates the exposure bias in traditional training but also improves the robustness and reliability of the model against unnatural errors. In addition to the training phase, we also utilize semantic information in the inference process. Traditional decoding methods assign equal or probability-based weights to all candidate outputs, which leads to up-voting more erroneous answers with higher co-occurrence. To address this problem, we design a novel decoding intervention strategy to maintain semantic consistency. This helps the language model to maintain semantic consistency in decoding and thus reduces the over-correction problem.

Our main contributions are summarized as follows:
(1) We implement a generative correction mechanism that enables models to correct multi-type errors.
(2) We introduce self-correcting adversarial training that derives adversarial examples from the model’s predictions, allowing the model to learn from its mistakes and effectively mitigate the exposure bias.
(3) To address the over-correction problem of language models, we design a novel decoding intervention strategy to maintain semantic consistency.
(4) Extensive experimental results on Chinese and English datasets show that our proposed method can correct multi-type errors or mistakes and can serve as a defense module in various natural language understanding and generation tasks.

\section{Related Work}

\subsection{Unnatural Text Correction}
Traditional Chinese text correction aims to address visual and phonetic errors caused by spelling errors. Early text correction methods adopted the process of recognizing and then correcting errors \cite{zhang2000automatic}. However, the effectiveness of these methods is limited by the varying accuracy of the identification and correction phases. To overcome these limitations, researchers have begun to explore end-to-end error correction methods. \citet{wang2019confusionset} utilized confusion sets and gating mechanisms, while \citet{zhang2020spelling} optimized detection and correction using BERT \cite{jin2020bert}. \citet{liu2021plome} introduced PLOME, which leverages a pre-trained masked language model that incorporates misspelling knowledge. \citet{li-etal-2022-learning-dictionary, li-etal-2022-past, li-etal-2022-improving-chinese} advanced text Correction techniques by learning heterogeneous knowledge from dictionaries, refining knowledge representations, and employing iterative correction strategies. \citet{wu-etal-2023-rethinking} improved language model performance through random masking, and \citet{Liu_Wu_Zhao_2024} rephrased sentences by filling slots.

Recently, \citet{feng-2024-protect} extended this task to non-natural text correction to address additional challenges facing Chinese text correction, such as errors arising from perfect pinyin, abbreviation pinyin, and character split.
However, perturbations in the form of insertions, deletions, inversions, and Unicode are still unexplored.

\begin{figure*}[t]
\centering
\includegraphics[width=0.9\textwidth]{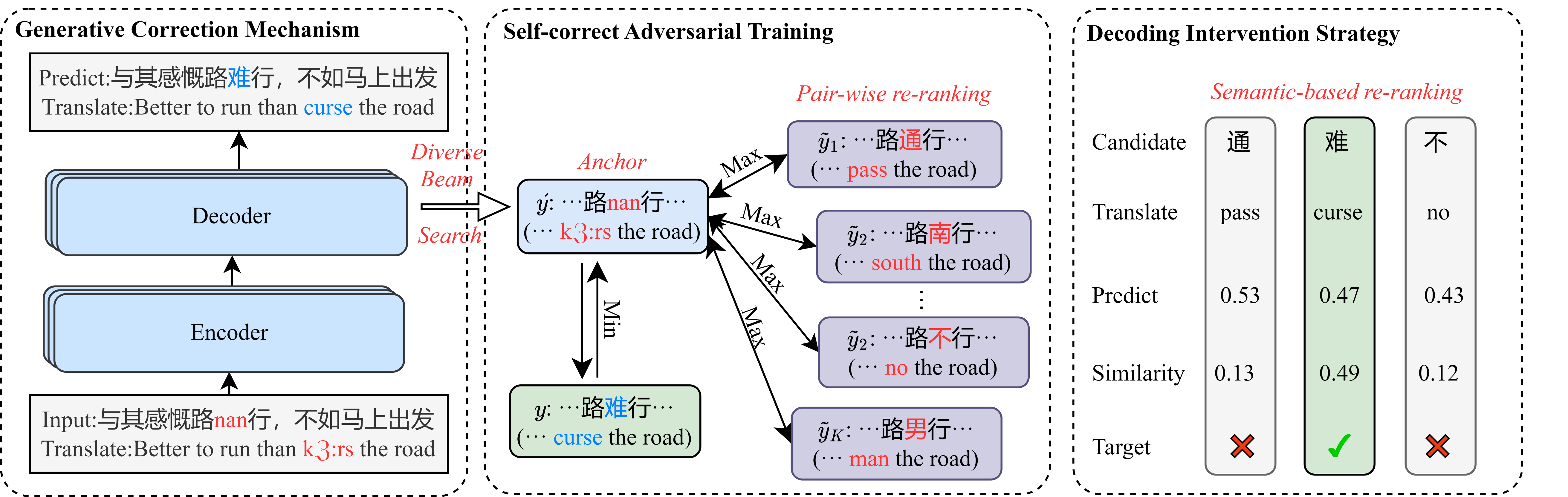}
\caption{The overall correction process of LIMIT. For easier understanding, pinyin errors in Chinese are represented by phonetic symbols in English. \label{fig:2}}
\end{figure*}

\subsection{Adversarial Training}
Adversarial Training (AT) \cite{goodfellow2014explaining} is a method used to improve a model’s defense against adversarial perturbations by training the model with adversarial examples, thereby enhancing its robustness against deceptive inputs. Most AT methods tend to defend against suboptimal adversarial examples that deceive the decoder \cite{Zhu2020FreeLB, jiang2020smart, aghajanyan2020better}. More recently, \citet{wu2023toward} proposed contextualized representation adversarial training to deviate the contextual representation of the encoder from potential adversarial influences. Additionally, \citet{gupta-etal-2023-dont} designed a text rewriting module to eliminate perturbations in the input.
Although these methods have made some progress in enhancing the model’s generalization ability, they can lead to significant performance degradation on the original task. Therefore, it is crucial to mitigate the exposure bias and improve the model’s adaptability and robustness to unseen scenarios.

\section{Method}
AT is usually applied to defend specific seen errors and perturbations. In contrast, real-world spelling errors and adversarial perturbations are subject to evolving model architectures and changing task contexts. To address the exposure bias problem associated with AT, as well as the inherent over-correction problem of language models, we propose a self-correct adversarial training framework for learning from mistakes (LIMIT), which consists of a generative correction mechanism, self-correct adversarial training, and a decoding intervention strategy. It corrects the text from unnatural errors through conditional generation. In this section, we illustrate the general design of the framework as well as the individual components. The overall process of our framework is shown in Figure \ref{fig:2}.

\subsection{Generative Correction Mechanism}
Mask-then-recovery is a commonly used correction mechanism for textual error correction. However, it fails to consider unseen multi-type errors and mistakes (such as pinyin, insertion, deletions, inversions, Unicode, character splitting, etc.) and unequal length errors (i.e., input and output lengths do not match) \cite{feng-2024-protect}. In contrast to the mask-then-recovery process, we propose a generative correction mechanism explicitly trained for eliminating spelling errors and adversarial perturbations.

Formally, given a clean text $\mathcal{X} = \{x_1, x_2, ..., x_n\}$, a bounded, imperceptible perturbation ${\delta}$ is added to produce an adversarial example $\mathcal{X}^{\prime} = \{x^{\prime}_1, x^{\prime}_2, ..., x^{\prime}_m\}$. Notably, the length $m$ of $\mathcal{X}^{\prime}$ possibly differs from the original input sentence $\mathcal{X}$, i.e., the length of the input sentence $\mathcal{X}^{\prime}$ is independent of the length of target sentence $\mathcal{Y}$. More specific perturbation processes can be found in the Technical Appendix. Traditional correction mechanisms can learn the optimal corrector by optimizing the following objectives:

\begin{equation}
\min _{f \in \mathcal{H}} \mathbb{E}_{(\mathcal{X}, \mathcal{Y}) \in \mathcal{D}} \max _{|\delta| \leq \epsilon} \ell[f(\mathcal{X}+\delta), \mathcal{Y}]
\end{equation}

Instead, our goal is to correct unnatural text errors and mistakes from the input and preserve the semantics of the original sentence. To this end, we implement a text-to-text generative mechanism, denoted by $\mathcal{P}_{r}$.

To effectively eliminate adversarial examples, the correct function $\mathcal{P}_{r}$ must recover the input $\mathcal{X}^{\prime}$ to the target $\mathcal{Y} = \{y_1, y_2, ..., y_n\}$:

\begin{equation}
\mathcal{Y}_i=\mathcal{P}_{r}\left(\mathcal{X}^{\prime}_i\right), \forall i \in\{1, n\}
\end{equation}
where the perturbed characters in $\mathcal{X^{\prime}}$ are replaced with the original ones to obtain $\mathcal{Y}$.

Compared to the mask-then-recovery process, the generative correction mechanism allows easier transfer to new error forms and tasks. This is a promising mechanism for correcting multi-type unnatural textual errors, in turn defending against potentially adversarial perturbations with word substitutions.

\subsection{Self-correct Adversarial Training}
The exposure bias problem in text correction occurs when a model is trained on a data distribution that does not accurately reflect the real-world scene. Unfortunately, real-world data distributions are more complex and variable, which may lead to poor generalization performance of the trained model.
To address this problem, we introduce self-correct adversarial training. It constructs adversarial examples from its own predictions via a beam search algorithm and implements the ranking loss to help calibrate robust representations.

Following the contrastive learning framework \cite{chen2020simple}, we train the model by comparing positive and negative sentence pairs to learn representations of ground truth sentences. By maximizing the similarity between source and target sequences while minimizing the similarity between negative sequences:

\begin{equation}
L^{\mathrm{NLL}}=-\log \frac{\exp \left(\operatorname{sim}\left(\boldsymbol{z}_{x^\prime}, \boldsymbol{z}_y\right) / \tau\right)}{\sum \exp \left(\operatorname{sim}\left(\boldsymbol{z}_{x^\prime}, \boldsymbol{z}_{y^\prime}\right) / \tau\right)}
\end{equation}
where $\boldsymbol{z}_{x^\prime}$, $\boldsymbol{z}_y$, $\boldsymbol{z}_{y\prime}$ denote vector representations of input $\mathcal{X^\prime}$, target $\mathcal{Y}$, and negative sample $\mathcal{Y}^\prime$, respectively. $\tau$ is the temperature, and $\operatorname{sim}(\cdot,\cdot )$ defines the cosine similarity.

However, training models using naive contrastive learning frameworks typically yield error corrections. In light of this, we propose a principled method for automatically constructing adversarial negative and positive examples that allow the model to fully utilize mistakes.
Specifically, we employ diverse beam search algorithms to dynamically create negative examples $\tilde{\mathcal{Y}}=\left\{\tilde{\mathcal{Y}}_1, \tilde{\mathcal{Y}}_2, \cdots, \tilde{\mathcal{Y}}_K\right\}$ from the top-$K$ list of model predictions. These self-generated negative examples are intended to enrich the generalization capability of the model by providing more realistic test-time predictions. 

We expect to fully utilize the model's mistakes, so we design a self-correcting loss function that realizes this property through pairwise comparisons. Specifically, we employ the sequence-level scores BLEU and similarity to quantify the generated examples. All examples were ranked according to their relative difference from the original sentence. Besides, the ranked example pairs are appended to the batch to form pairwise examples $(\tilde{\mathcal{Y}}_1^{+}, \tilde{\mathcal{Y}}_2^{-})$, where $+$ and $-$ are determined by their ranks.
We optimize the model parameters using the weighted sum of the negative log-likelihood loss $L^\mathrm{NLL}$ and the self-correct ranking loss $L^\mathrm{RANK}$ as the training loss for each training pair $(\mathcal{X}^{\prime}, \mathcal{Y})$ as follows:

\begin{equation}
\begin{gathered}
=\\\sum_{a \notin \mathcal{Y}} \sum_{b \in \mathcal{Y}} \max \left(0, \gamma+\operatorname{sim}\left(z_{\mathcal{X}^{\prime}},\tilde{z_{\mathcal{Y}_a^{+}}}\right),\operatorname{sim}\left(z_{\mathcal{X}^{\prime}},\tilde{z_{\mathcal{Y}_b^{+}}}\right)\right)\\
L=L^{\mathrm{NLL}}+L^{\mathrm{RANK}} \\
\end{gathered}
\end{equation}
During training, the $L^\mathrm{NLL}$ increases the similarity between the model output $\mathcal{X}^{\prime}$ and the target sentence $\mathcal{Y}$. The $L^\mathrm{RANK}$ prevents the model from generating each counter-example containing an adversarial perturbation $\tilde{\mathcal{Y}}_k$, $\gamma$ is the margin.

\subsection{Decoding Intervention Strategy}

The semantics of discrete text may be affected by even subtle errors and perturbations. Traditional decoding strategies may lead to a dramatic performance degradation under adversarial perturbations. Therefore, we design a decoding intervention strategy to address the over-correction problem and further improve the model's robustness. Specifically, we incorporate a similarity function into the decoding phase to dynamically evaluate the correctness of the next token predicted by the decoder. The decoding goal in LIMIT is to find the sequence $\mathcal{Y}$ that maximizes the likelihood of the learned similarity score and the regular language model:

\begin{equation}
\begin{aligned}
s(\mathcal{X}^{\prime}, \mathcal{Y})=\sum_{t=1}^{|\mathcal{Y}|} & \left(\log p_\theta\left(\mathcal{Y}_t \mid \mathcal{Y}_{<t}, \mathcal{X}^{\prime}\right)\right. \\
& \left.+\alpha \times \operatorname{sim}\left(\mathcal{Y}_t, \mathcal{X}^{\prime}\right)\right)
\end{aligned}
\end{equation}
where the first term is the original probability of the language model, and the second term is the similarity score between the given the input sentence $\mathcal{X}^{\prime}$ and the generated ${\mathcal{Y}_{t}}$, and $\alpha$ is the hyper-parameter that balances the contribution of each term.

\begin{table*}[]
\centering
\begin{tabular}{l|cc|cc|cc|cc|cc}
\toprule
\multicolumn{1}{c|}{\multirow{2}{*}{\textbf{Model}}} & \multicolumn{2}{c|}{\textbf{Perfect Pinyin}} & \multicolumn{2}{c|}{\textbf{Abb. Pinyin}} & \multicolumn{2}{c|}{\textbf{Char. Split}} & \multicolumn{2}{c|}{\textbf{Hybrid}} & \multicolumn{2}{c}{\textbf{Hybrid-v2}} \\ \cmidrule{2-11} 
\multicolumn{1}{c|}{}                                & Pre                   & F1                   & Pre                     & F1                      & Pre                   & F1                    & Pre               & F1               & Pre              & F1              \\ \midrule

BERT                                                 & 31.0                  & 33.0                 & 41.0                    & 43.0                    & 43.4                  & 54.4                  & 25.7              & 28.3             & 14.6                    & 15.8                   \\
SoftMasked                                           & 32.0                  & 34.5                 & 45.7                    & 46.8                    & 43.4                  & 50.6                  & 22.0              & 25.8             & 15.3                    & 16.7                   \\
MDCSpell                                             & 32.6                  & 34.6                 & 45.5                    & 46.5                    & 42.3                  & 49.8                  & 23.1              & 27.0             & 14.3                    & 15.5                   \\
PLOME                                                & 59.5                  & 59.7                 & 33.5                    & 35.1                    & 2.9                   & 3.2                   & 48.9              & 48.8             & -                       & -                      \\
MFT                                                  & 30.8                  & 32.8                 & 30.2                    & 31.7                    & 39.6                  & 46.8                  & 48.0              & 56.2             & -                       & -                      \\
ReLM                                                 & -                     & -                    & 46.7                    & 47.9                    & -                     & -                     & 37.0              & 43.5             & -                       & -                      \\ \midrule
RobustGEC                                            & 58.6                  & 53.2                 & 46.9                    & 30.2                    & 67.7                  & 65.5                  & 51.7              & 38.5             & 16.4                        & 15.4                       \\
ATINTER                                              & 70.0                  & 61.2                 & 58.0                    & 35.6                    & 82.0                  & 78.7                  & 59.1              & 41.3             & 54.1                        & 23.5                       \\
PROTECT-Fewshot                                      & \underline{73.4}            & 67.0                 & 68.7                    & 45.4                    & 81.4                  & 78.7                  & 66.8              & 47.4             & 77.1                    & 50.3                   \\
PROTECT-Finetune                                     & \textbf{90.2}         & \underline{82.1}           & \textbf{84.8}           & \underline{57.7}              & \textbf{94.4}         & \underline{91.8}            & \underline{90.4}        & \underline{71.2}       & \underline{83.1}                        & \underline{59.6}                       \\
LIMIT(Ours)                                          & $\textbf{90.2}^\dagger$         & $\textbf{84.6}^\dagger$        & \underline{69.8}              & $\textbf{63.5}^\dagger$           & \underline{91.9}            & $\textbf{93.2}^\dagger$         & $\textbf{91.6}^\dagger$     & $\textbf{81.2}^\dagger$    & $\textbf{84.8}^\dagger$                        & $\textbf{66.8}^\dagger$                       \\ \hline
GPT-3.5-Turbo-10shot                                 & 23.2                  & 22.1                 & 2.7                     & 3.4                     & 1.0                   & 1.2              & 22.2                     & 19.4     & 12.5               & 11.0                                  \\ \bottomrule
\end{tabular}%
\caption{Performance of the baseline model and our approach on five Chinese unnatural text correction datasets. The best and second-best results are highlighted in \textbf{bold} and \underline{underline}. Where Abb.  Pinyin and Char. Split represents the Abbreviation Pinyin and Character Split respectively. The superscript $\dagger$ indicates p $<$ 0.05 for the t-test of the LIMIT vs. the PROTECT-Finetune.\label{tab:1}} 
\end{table*}

\section{Experiments}

In this section, we compare LIMIT with a range of text correction methods on unnatural text correction datasets. We also evaluate the adoption of LIMIT as a defense method against perturbations on natural language generation (NLG) tasks and natural language understanding (NLU) compared to adversarial training methods.

\subsection{Datasets}
\textbf{Unnatural Text Correction Datasets}:
\textit{PROTECT} \cite{feng-2024-protect} includes unnatural text errors that are possible in Chinese characters. There are four subdatasets \textit{Perfect Pinyin}, \textit{Abbreviation Pinyin}, \textit{Character Split}, and \textit{Hybrid}. It covers common spelling errors involving visually or phonetically similar characters, splitting characters into radicals, and converting characters to perfect or abbreviated pinyin forms. \textit{Hybrid-v2} Based on the Hybrid perturbations, we construct insertion, deletion, inversion, and Unicode perturbations.

To verify that the proposed method can effectively serve as an adversarial defense method, following \cite{su2022rocbert} and \cite{feng-2024-protect}, we perturb the NLU and NLG datasets. The specific perturbation process is detailed in the Technical Appendix.

\textbf{NLU Datasets}:
For Chinese datasets, \textit{TNEWS} \cite{xu-etal-2020-clue} is a Chinese dataset for text classification. \textit{AFQMC} \cite{xu-etal-2020-clue} is a Chinese question-matching dataset designed to evaluate the performance of natural language processing models. \textit{CMNLI} \cite{xu-etal-2020-clue} is a Chinese multi-genre cross-domain natural language reasoning dataset that asses a model's ability to determine the relationships between premises and hypotheses. \textit{IFLYTEK} \cite{xu-etal-2020-clue} is a Chinese long-text classification dataset. \textit{COLD} \cite{deng-etal-2022-cold} is a Chinese offensive speech detection dataset.

For English datasets, we conduct our experiments on \textit{advSST-2}, \textit{advQQP}, \textit{advMNLI}, and \textit{advRTE} \cite{wang2021adversarial}, which applies 14 state-of-the-art textual adversarial attack methods to GLUE tasks.

\textbf{NLG Datasets}:
\textit{ADGEN} \cite{shao-etal-2019-long} is an advertisement generation dataset. \textit{CSL} \cite{zhang-etal-2021-ambert} is an academic domain text summarization dataset consisting of abstracts and titles of publications in the field of computer science. \textit{LCSTS} \cite{hu-etal-2015-lcsts} is a large Chinese short text summarization dataset.

\subsection{Baselines}

\textbf{Text Correct Baselines}: 
\textit{BERT} \cite{devlin-etal-2019-bert} is a pre-trained language model that can be used for fine-tuning various natural language processing tasks.
\textit{SoftMasked} \cite{zhang-etal-2020-spelling} used a pipeline structure of detection network and correction network to implement text error correction.
\textit{MDCSpell} \cite{zhu-etal-2022-mdcspell} employed a late fusion strategy to integrate the hidden states of the corrector with those of the detector, aiming to mitigate the adverse effects caused by misspelled characters.
\textit{PLOME} \cite{liu2021plome} designed a masking strategy based on a semantic confusion set when training pre-trained language models.
\textit{MFT} \cite{wu-etal-2023-rethinking} randomly masked 20\% of the non-error tokens in the input sequence during the fine-tuning process, which is enough to learn a better language model without sacrificing the error model.
\textit{RobustGEC} \cite{zhang-etal-2023-robustgec} proposed an effective post-training method Context Perturbation Robustness to enhance the stability and reliability of these systems in real-world applications.
\textit{ATINTER} \cite{gupta-etal-2023-dont} is a module that intercepts and learns to rewrite adversarial inputs, making it non-adversarial for downstream text classifiers.
\textit{ReLM} \cite{Liu_Wu_Zhao_2024} trained the model to restate entire sentences by filling in extra slots instead of marking them word by word.

\textbf{Adversarial Training Baselines}: 
\textit{FreeLB} \cite{Zhu2020FreeLB} is a fast adversarial training algorithm that integrates each intermediate example into a backward pass.
\textit{SMART} \cite{jiang2020smart} introduced smoothness-induced regularization in adversarial training for better generalization performance.
\textit{R3F} \cite{aghajanyan2020better} replaced the previously used adversarial targets with parametric noise (sampled from a normal or uniform distribution).
\textit{CreAT} \cite{wu2023toward} presented a simple and effective contextual representation-adversarial training, where the attack is to explicitly optimize the contextual representation of the deviation encoder.
\textit{Match-Tuning} \cite{ijcai2022p0610} added regularization between examples in the same batch.

\textbf{Large Language Models}:
\textit{Llama}\footnote{https://github.com/LlamaFamily/Llama-Chinese}, \textit{Baichuan2} \cite{baichuan2023baichuan2}, \textit{OPT-66B} \cite{zhang2022opt}, \textit{BLOOM} \cite{le2023bloom} and \textit{ChatGPT}. More experimental results for large language models are presented in the Technical Appendix.

\subsection{Implementations}

To obtain robust textual representations against unnatural textual errors. For the Chinese corpus, we constructed adversarial examples using 300k randomly extracted texts from Chinese Wikipedia and continued pre-training on the T5-Base-Chinese \footnote{https://huggingface.co/uer/t5-base-chinese-cluecorpussmall} model. Similarly, for the English corpus, we constructed adversarial examples using 300k randomly extracted texts from Comments2019. Likewise, we continued pre-training on the T5-Large\footnote{https://huggingface.co/google-t5/t5-large} model.

To obtain robust representations, we pretrained the generation model after constructing adversarial examples. LIMIT has 12 layers/heads and 768 hidden neurons. It undergoes training on a scale of 60k with a batch size of 32, a learning rate of 1e-5, and a warm-up stage of 6k. The English version consists of 48 layers, 24 attention heads, and 1024 hidden neurons. It follows a learning rate of 1e-5, a warm-up stage of 6k, a batch size of 32, and a training stage of 60k.

LIMIT introduces three additional hyperparameters. The first one is the diversity of beam search size, denoted as $K$. The second one is the boundary strength, denoted as $\gamma$. The third one is the balancing factor, denoted as $\alpha$. For all datasets, we set $K$ to 12 and $\gamma$ to 0.01. We tune $\alpha$ on the validation set using values from [0.3, 0.4, 0.5, 0.6, 0.7]. In practice, increasing the number of dynamic negative samples continually improves performance.

For the unnatural text correction task, we evaluate performance using precision (Pre) and the F1 score. In the NLU task, accuracy (Acc) serves as our primary metric. For NLG tasks, we employ Rouge-1 (R-1), Rouge-2 (R-2), and Rouge-L (R-L) to assess the quality of the generated text in comparison to the target text. These three metrics provide different perspectives on the quality of the generated text.

\begin{table*}[t!]
\centering
\begin{tabular}{l|cc|cc|cc|cc|cc}
\toprule
\multicolumn{1}{c|}{\multirow{3}{*}{\textbf{Model}}} &
  \multicolumn{2}{c|}{\textbf{TNEWS}} &
  \multicolumn{2}{c|}{\textbf{AFQMC}} &
  \multicolumn{2}{c|}{\textbf{CMNLI}} &
  \multicolumn{2}{c|}{\textbf{IFLYTEK}} &
  \multicolumn{2}{c}{\textbf{COLD}} \\ \cmidrule{2-11} 
\multicolumn{1}{c|}{} &
  Clean &
  Adv &
  Clean &
  Adv &
  Clean &
  Adv &
  Clean &
  Adv &
  Clean &
  Adv \\
\multicolumn{1}{c|}{} &
  (Acc) &
  (Acc) &
  (Acc) &
  (Acc) &
  (Acc) &
  (Acc) &
  (Acc) &
  (Acc) &
  (Acc) &
  (Acc) \\ \hline
BERT &
  66.6 &
  65.4 &
  75.1 &
  72.4 &
  80.8 &
  77.5 &
  58.4 &
  56.2 &
  93.1 &
  80.5 \\
FreeLB &
  67.1 &
  65.5 &
  74.2 &
  70.9 &
  80.1 &
  77.4 &
  59.3 &
  57.6 &
  93.1 &
  80.5 \\
SMART &
  66.6 &
  64.7 &
  73.1 &
  70.9 &
  79.4 &
  76.3 &
  58.3 &
  55.5 &
  93.1 &
  80.5 \\
R3F &
  67.1 &
  65.5 &
  74.1 &
  71.0 &
  80.1 &
  77.5 &
  58.7 &
  56.5 &
  93.1 &
  80.5 \\
CreAT &
  66.8 &
  65.4 &
  73.4 &
  70.5 &
  79.0 &
  76.0 &
  58.9 &
  57.2 &
  93.1 &
  80.5 \\
BERT+LIMIT(Ours) &
  66.6 &
  \textbf{66.0} &
  75.1 &
  \textbf{72.4} &
  80.8 &
  \textbf{79.1} &
  58.4 &
  \textbf{59.7} &
  93.1 &
  \textbf{82.4} \\ \hline
Llama-7B &
  13.0 &
  10.7 &
  43.5 &
  49.1 &
  34.9 &
  34.9 &
  47.6 &
  48.7 &
  50.0 &
  43.8 \\
Baichuan2-13B &
  33.2 &
  27.9 &
  69.0 &
  69.0 &
  34.4 &
  33.5 &
  44.8 &
  44.0 &
  48.2 &
  47.5 \\
GPT-3.5-Turbo-10shot &
  49.9 &
  47.9 &
  69.0 &
  68.9 &
  52.9 &
  51.0 &
  49.8 &
  37.1 &
  51.9 &
  50.0 \\ \hline
\end{tabular}%
\caption{\label{tab:2}
Performance of the adversarial training baseline models and our method on the Chinese NLU dataset. The best results are labeled with \textbf{bold}.
}
\end{table*}

\begin{table}[t]
\renewcommand\arraystretch{1.0}
\resizebox{\columnwidth}{!}{%
\begin{tabular}{lcccc}
\toprule
\multicolumn{1}{c|}{\multirow{2}{*}{\textbf{Model}}} & \multicolumn{1}{c|}{\textbf{\begin{tabular}[c]{@{}c@{}}Adv\\ SST-2\end{tabular}}} & \multicolumn{1}{c|}{\textbf{\begin{tabular}[c]{@{}c@{}}Adv\\ QQP\end{tabular}}} & \multicolumn{1}{c|}{\textbf{\begin{tabular}[c]{@{}c@{}}Adv\\ MNLI-m\end{tabular}}} & \textbf{\begin{tabular}[c]{@{}c@{}}Adv\\ RTE\end{tabular}} \\
\multicolumn{1}{c|}{}                                & \multicolumn{1}{c|}{(Acc)}                                                        & \multicolumn{1}{c|}{(Acc)}                                                      & \multicolumn{1}{c|}{(Acc)}                                                         & (Acc)                                                      \\ \midrule
\multicolumn{1}{l|}{BERT $\natural $}                            & \multicolumn{1}{c|}{32.3}                                                         & \multicolumn{1}{c|}{50.8}                                                       & \multicolumn{1}{c|}{32.6}                                                          & 37.0                                                       \\
\multicolumn{1}{l|}{FreeLB $\natural $}                         & \multicolumn{1}{c|}{31.6}                                                         & \multicolumn{1}{c|}{51.0}                                                       & \multicolumn{1}{c|}{33.5}                                                          & 42.0                                                       \\
\multicolumn{1}{l|}{R3F $\natural $}                             & \multicolumn{1}{c|}{38.5}                                                         & \multicolumn{1}{c|}{40.6}                                                       & \multicolumn{1}{c|}{35.8}                                                          & 50.1                                                       \\
\multicolumn{1}{l|}{CreAT $\natural $}                           & \multicolumn{1}{c|}{35.3}                                                         & \multicolumn{1}{c|}{51.5}                                                       & \multicolumn{1}{c|}{36.0}                                                          & 45.2                                                       \\
\multicolumn{1}{l|}{Match-Tuning $\natural $}                    & \multicolumn{1}{c|}{51.4}                                                         & \multicolumn{1}{c|}{41.5}                                                       & \multicolumn{1}{c|}{35.5}                                                          & 47.5                                                       \\
\multicolumn{1}{l|}{BERT+LIMIT(Ours)}                           & \multicolumn{1}{c|}{\textbf{66.2}}                                                & \multicolumn{1}{c|}{\textbf{78.8}}                                                       & \multicolumn{1}{c|}{\textbf{69.4}}                                                 & \textbf{84.0}                                              \\ \midrule
\multicolumn{1}{l|}{OPT-66B $\flat$}                       & \multicolumn{1}{c|}{52.4}                                                         & \multicolumn{1}{c|}{46.1}                                                       & \multicolumn{1}{c|}{39.7}                                                          & 42.0                                                       \\
\multicolumn{1}{l|}{BLOOM-176B $\flat$}                    & \multicolumn{1}{c|}{51.3}                                                         & \multicolumn{1}{c|}{41.0}                                                       & \multicolumn{1}{c|}{26.4}                                                          & 43.2                                                       \\
\multicolumn{1}{l|}{GPT-3.5-Turbo-10shot}            & \multicolumn{1}{c|}{60.1}                                                         & \multicolumn{1}{c|}{72.0}                                              & \multicolumn{1}{c|}{67.8}                                                          & 75.3                                                       \\ \bottomrule
\end{tabular}%
}
\caption{Performance of the adversarial training baseline models and our method on the English AdvGLUE dataset. Partial experimental results are from \cite{wu2023toward}$\natural $ and \cite{wang2023robustness}$\flat$, with the best performing scores shown in \textbf{bold}. More results for the large language model are presented in the Technical Appendix. \label{tab:3}}
\end{table}

\subsection{Results on Chinese Unnatural Text Correction Datasets}

In the unnatural text correction task, we evaluate the performance of several baseline models and our proposed LIMIT in five different types of unnatural text correction tasks: perfect pinyin, abbreviation pinyin, character split, hybrid, and hybrid-v2. The experimental results are shown in Table \ref{tab:1}. To guarantee the reliability of the experiments, all results are averaged over five experiments.

We analyze the performance of traditional Bert-based text correction methods. PLOME performed best in the perfect pinyin task with an F1 score of 59.7\%, but worst in the character split task with an F1 score of only 3.2\%. ReLM performed well in the abbreviation pinyin task with an F1 score of 47.9\% but failed to correct the errors in the other tasks. Overall, traditional text correction methods perform poorly in unnatural text correction tasks with generally low F1 scores. For the harder Hybrid-v2 dataset, PLOME, MFT, and ReLM cannot handle such errors.

Furthermore, we analyze the performance of the generative correction methods. RobustGEC is the first to consider the robustness of a text error correction task against perturbations, however, the method performs poorly on the unnatural text correction tasks. The text rewriting strategy adopted by ATINTER performed well in the perfect pinyin and Character split tasks, with F1 scores of 61.2\% and 78.7\%, respectively, but did not perform well in the more complex tasks. The PROTECT-Fewshot model performed well on the perfect pinyin, abbreviation pinyin, and character split tasks with F1 scores of 67.0\%, 45.4\%, and 78.7\% respectively, but performed slightly poorer on the hybrid task with F1 scores of 61.2\% and 78.7\% respectively. This is because the method excels at predicting error accuracy but is less effective at correcting errors.

It is noteworthy that our proposed LIMIT model demonstrates exceptional performance across all tasks, while the large language model is ineffective at correcting errors in unnatural text. This suggests that the LIMIT has significant advantages and potential in the Chinese unnatural text correction task.

\subsection{Results on Chinese NLU Datasets}

Table \ref{tab:2} shows the experimental results on five Chinese NLU datasets. The experimental demonstrates show that the adversarial robustness of our proposed LIMIT achieves consistent improvement on the NLU task. On the perturbed TNEWS, AFQMC, CMNLI, IFLYTEK, and COLD datasets, LIMIT obtains an improvement of 0.6\%, 1.9\%, 3.1\%, 2.5\%, and 1.9\%, respectively. We find that all the adversarial training methods suffer a loss of performance in Chinese tasks. The trade-off between performance and robustness is consistent with previous findings. For LIMIT, however, it is only responsible for removing adversarial perturbations from the input text. This preserves the performance of the language model to some extent.
For example, the FreeLB and R3F outperform vanilla BERT on clean TNEWS and IFLYTEK datasets, while the SMART and CreAT sacrifice prediction performance on all tasks. On the relatively easier classification adversarial datasets TNEWS and IFLYTEK, most of the methods provide performance gains. However, for the inference tasks AFQMC and CMNLI, both lead to performance loss when trading off performance and robustness. 

\subsection{Results on English NLU Datasets}

\begin{table}[]
\resizebox{\columnwidth}{!}{%
\begin{tabular}{lcccccc}
\toprule
\multicolumn{1}{c|}{\multirow{2}{*}{\textbf{Model}}} & \multicolumn{3}{c|}{\textbf{Clean}} & \multicolumn{3}{c}{\textbf{Adv}} \\
\multicolumn{1}{l|}{}          & (R-1) & (R-2) & \multicolumn{1}{c|}{(R-L)} & (R-1) & (R-2) & (R-L) \\ \midrule
\multicolumn{7}{c}{\textbf{ADGEN}}                                                                  \\ \midrule
\multicolumn{1}{l|}{RobustGEC} & 43.9  & 18.9  & \multicolumn{1}{c|}{26.8}  & 40.6  & 15.6  & 23.9  \\
\multicolumn{1}{l|}{PROTECT}   & 42.7  & 18.9  & \multicolumn{1}{c|}{27.3}  & 39.0  & 15.4  & 24.1  \\
\multicolumn{1}{l|}{ATINTER}   & 43.9  & 18.9  & \multicolumn{1}{c|}{26.8}  & 37.8  & 14.1  & 23.6  \\
\multicolumn{1}{l|}{LIMIT(Ours)}     & 43.9  & 18.9  & \multicolumn{1}{c|}{26.8}  & \textbf{41.6}  & \textbf{16.7}  & \textbf{24.9}  \\ \midrule
\multicolumn{7}{c}{\textbf{CSL}}                                                                    \\ \midrule
\multicolumn{1}{l|}{RobustGEC} & 64.6  & 52.6  & \multicolumn{1}{c|}{61.4}  & 52.9  & 38.2  & 49.9  \\
\multicolumn{1}{l|}{PROTECT}   & 63.6  & 52.0  & \multicolumn{1}{c|}{60.7}  & 52.8  & 37.7  & 49.0  \\
\multicolumn{1}{l|}{ATINTER}   & 64.6  & 52.6  & \multicolumn{1}{c|}{61.4}  & 48.8  & 35.3  & 46.1  \\
\multicolumn{1}{l|}{LIMIT(Ours)}     & 64.6  & 52.6  & \multicolumn{1}{c|}{61.4}  & \textbf{58.9}  & \textbf{45.3}  & \textbf{55.5}  \\ \midrule
\multicolumn{7}{c}{\textbf{LCSTS}}                                                                  \\ \midrule
\multicolumn{1}{l|}{RobustGEC} & 44.0  & 29.3  & \multicolumn{1}{c|}{40.7}  & 35.0  & 21.0  & 32.4  \\
\multicolumn{1}{l|}{PROTECT}   & 42.0  & 27.4  & \multicolumn{1}{c|}{38.9}  & 35.1  & 21.1  & 32.6  \\
\multicolumn{1}{l|}{ATINTER}   & 44.0  & 29.3  & \multicolumn{1}{c|}{40.7}  & 39.3  & 24.5  & 36.0  \\
\multicolumn{1}{l|}{LIMIT(Ours)}     & 44.0  & 29.3  & \multicolumn{1}{c|}{40.7}  & \textbf{39.4}  & \textbf{24.6}  & \textbf{36.1}  \\ \bottomrule
\end{tabular}%
}
\caption{Performance of the generative correction baseline models and our method on the NLG dataset. The best results are labeled with \textbf{bold}. \label{tab:4}}
\end{table}

\begin{table}[h!]
\resizebox{\columnwidth}{!}{%
\begin{tabular}{c|c|c|ccc}
\toprule
\multirow{2}{*}{\textbf{\begin{tabular}[c]{@{}c@{}}Dataset\\ Model\end{tabular}}} & \textbf{Hybrid} & \textbf{CMNLI} & \multicolumn{3}{c}{\textbf{CSL}}              \\
                                 & (F1) & (Acc) & (R-1) & (R-2) & (R-L) \\ \midrule
\multicolumn{1}{l|}{Fine-tuning} & 75.2 & 77.5  & 52.8  & 38.2  & 49.9  \\
\multicolumn{1}{r|}{+SC}         & 75.6 & 78.8  & 58.0  & 45.1  & 55.4  \\
\multicolumn{1}{r|}{+DI}                                                          & \textbf{81.2}   & \textbf{79.1}  & \textbf{58.9} & \textbf{45.3} & \textbf{55.5} \\ \bottomrule
\end{tabular}%
}
\caption{Ablation results in different components of LIMIT. The best-performing scores are in bold. Results for additional datasets are provided in the Technical Appendix.}
\label{tab:5}
\end{table}

Table \ref{tab:3} shows the experimental results on four English NLU adversarial datasets (AdvGLUE). Experimental results illustrate that LIMIT outperforms state-of-the-art methods and achieves the best performance on four randomly selected datasets.
For the pre-training and fine-tuning methods, Match-Tuning with BERT-large achieves competitive results by adding regularization. For the large language models, ChatGPT exhibits better performance than the specifically designed model, achieving accuracy scores of 60.1\%, 72.0\%, 67.8\%, and 65.5\% on the four datasets, respectively. However, models with the same parameter sizes show considerable variation in performance, with an average accuracy of only 42.2\% for BLOOM. For adversarial training methods, FreeLB, R3F, and CreAT perform poorly, which has validated their struggles to cope with a multitype of errors and perturbations.

\subsection{Results on Chinese NLG Datasets}

The experimental results on the Chinese NLG dataset are demonstrated in Table \ref{tab:4}. The results show that LIMIT exhibits the best adversarial robustness. This reflects its transferability to new tasks and models with competitive performance. For the experimental results with BobustGEC as the backbone, there is an average improvement of 1.0\%, 6.2\%, and 3.9\% for ADGEN, CSL, and LCSTS adversarial datasets at Rouge-1, Rouge-2, and Rouge-L, respectively.
Although PROTECT is designed to correct unnatural errors, it performs poorly on all three adversarial perturbed datasets. While the ATINTER, which employs rewriting to mitigate adversarial perturbations, incurs a performance loss on the ADGEN and CSL adversarial datasets. Specifically, on the ADGEN adversarial dataset, it degrades by 2.8\%, 1.5\%, and 0.3\% on Rouge-1, Rouge-2, and Rouge-L, respectively. Likewise, it drops by 4.1\%, 2.9\%, and 3.8\% in the CSL dataset, respectively.

\subsection{Ablation Study}

\begin{figure}[t!]
\centering
\includegraphics[width=\columnwidth]{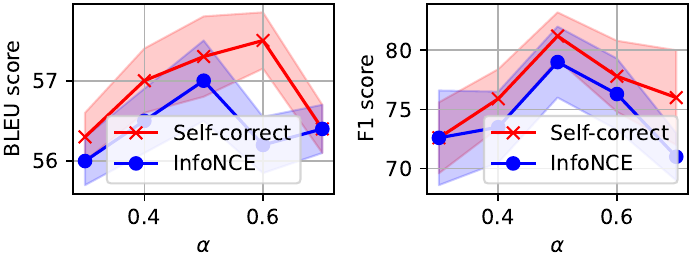}
\caption{Relationship between $\alpha$ and BLEU under different training losses on the Chinese unnatural text correction dataset (Hybrid).}
\label{fig:3}
\end{figure}

\begin{table}[t!]
\centering
\resizebox{\columnwidth}{!}{%
\begin{tabular}{l|cccc}
\toprule
\multirow{2}{*}{\textbf{Task}}   & \textbf{Pinyin} & \textbf{Abb.}    & \textbf{Char.} & \textbf{Hybrid} \\
                & \multicolumn{4}{c}{\#Overcorrections / \#Undercorrections}     \\ \midrule
Vanilla         & 151/636         & 107/428         & 131/286         & 124/522         \\
PROTECT         & 92/423          & \textbf{22}/319 & 26/72  & 48/310 \\ 
LIMIT(Ours)         & \textbf{43/149}          & 31/\textbf{310} & \textbf{24/67}  & \textbf{33/134} \\ \bottomrule
\end{tabular}%
}
\caption{The empirical analysis results of the Vanilla fine-tune, PROTECT, and the proposed method in this paper. Where Abb. and Char. are Abbreviation Pinyin and Character Split respectively. Following \citet{feng-2024-protect}, we statistically counted the quantity of overcorrected samples.}
\label{tab:6}
\end{table}

Table \ref{tab:5} shows the ablation studies of the different components of LIMIT on the Chinese and English adversarial datasets. It indicates that the components of self-correct adversarial training (SC) and decoding intervention (DI) both play key roles in enhancing adversarial robustness. Specifically, with the addition of SC, the average accuracy of the NLU dataset is improved by an average of 2.2\%, and the Rouge-L of the NLG dataset is improved by an average of 3.1\%. Similarly, with the addition of DI, the average accuracy of the NLU dataset is improved by an average of 0.4\%, and an average of 0.3\% improves the Rouge-L of the NLG dataset.

\subsection{Empirical Analysis on Hyper-parameter}

Figure \ref{fig:3} shows the impact of the parameter $\alpha$ on BLEU scores and accuracy under different training losses. The proposed self-correct ranking loss achieves the best performance at $\alpha = 0.5$, with a BLEU score of 0.57 and an F1 score of 81.2\%. In comparison, the traditional adversarial training loss, InfoNCE, reaches a BLEU score of 0.56 and an accuracy of 79.3\% at $\alpha = 0.5$. It demonstrated the effectiveness of the self-correct ranking loss in Chinese unnatural text correction.

\subsection{Empirical Analysis on Over-correction}

LIMIT achieves the best performance in perfect pinyin, character split, and hybrid, as shown in Table \ref{tab:6}. Nevertheless, improving correction accuracy for abbreviated pinyin remains a necessary direction that requires further effort.

\section{Conclusion}
In this paper, we propose a self-correct adversarial training framework for learning from mistakes, LIMIT, that enhances model robustness and adaptability to evolving spelling errors and adversarial perturbations. LIMIT offers a model- and task-agnostic solution for correcting unnatural text errors, ensuring robustness in error correction. Furthermore, it showcases transferability to various natural language understanding and natural language generation tasks, effectively resisting multi-type errors and perturbations.

\section{Acknowledgments}
	This work was supported by the National Natural Science Foundation of China (Grant No. U22A2099 and Grant No. 62336003).

\section{Technical Appendix}

\subsection{Adversarial Examples Construction}
To bridge the gap between evolving spelling errors, adversarial perturbations, and real-world adversarial examples, we construct adversarial examples by using word substitution-based perturbations and noise injection.

For Chinese text, following \cite{su2022rocbert, liu2023twins}, all possible perturbations that can occur in Chinese tokens are taken into account:

\textbf{\textit{Typos}}: Common spelling errors replace characters with visually similar characters\footnote{http://kanji-database.sourceforge.net} or phonetically similar characters\footnote{https://unicode.org/charts/unihan.html}. There is also the practice of splitting a character into multiple characters based on radicals and replacing them with the split characters. 

\textbf{\textit{Pinyin}}: Converts a Chinese character into Pinyin (perfect pinyin) or abbreviation pinyin (the initial form of pinyin) form.

\textbf{\textit{General}}: Randomly sample a visually similar Unicode as a replacement. Sample a character from a vocabulary set and randomly insert that character to the left or right of the current character. Swap characters with their neighbors. Deletes the character directly.

For English text, following \cite{morris2020textattack, wang2021adversarial}, we adopt the following perturbation approaches:

\textbf{\textit{Wordnet}}: augments text by replacing words with WordNet synonyms.

\textbf{\textit{Embedding}}: augments text by replacing words with neighbors in the counter-fitted embedding space, with a constraint to ensure their cosine similarity is at least 0.8.

\textbf{\textit{Charswap}}: augments text by substituting, deleting, inserting, and swapping adjacent characters.

\textbf{\textit{Eda}}: augments text with a combination of word insertions, substitutions, and deletions.

\textbf{\textit{Checklist}}: augments text by contraction/extension and by substituting names, locations, and numbers.

\textbf{\textit{Clare}}: augments text by replacing, inserting, and merging with a pre-trained masked language model.

\textbf{\textit{Back-trans}}: augments text by back-translation approach.

Accordingly, the process of adversarial perturbing can be formalized as follows:
\begin{equation}
\begin{aligned}
\delta & =\min \left(\max (\operatorname{int}(\epsilon), 1), e\right) \\
\epsilon & \sim \mathcal{N}\left(\max \left(1,0.15 e\right), 1\right)
\end{aligned}
\end{equation}
where $\delta$ is the mask ratio, following BERT \cite{jin2020bert} we mask 15\% of the characters, $e$ is the number of characters in the sentence, and the int(·) function rounds to the nearest integer. If the sentence is too short, we will make sure to attack at least one character.

\begin{table}[t!]
\centering
\begin{tabular}{c|cccc}
\toprule
\textbf{Task}   & \textbf{Train} & \textbf{Test} & \textbf{\#Pos} & \textbf{\#Neg} \\ \midrule
Perfect Pinyin  & 23,502    & 1,000   & 499  & 501 \\
Abbreviation Pinyin        & 26,910  & 1,000 & 499 & 501 \\
Character Split & 25,306    & 1,000   & 488  & 512 \\
Hybrid          & 24,839    & 1,000   & 495  & 505 \\
Hybrid-v2       & 26,766    & 1,000   & 487  & 513 \\\bottomrule
\end{tabular}
\caption{Statistics for different datasets. \label{tab:7}}
\end{table}

For Chinese natural language understanding and generation datasets, we apply the aforementioned Chinese perturbation techniques to the test sets while keeping the training sets unchanged. This approach ensures that the test data is subjected to the same adversarial conditions, allowing us to evaluate how well the models handle these disturbances without altering the training environment.

For English natural language understanding datasets, we validate our methods using the AdvGLUE \cite{wang2021adversarial} dataset. This dataset is specifically designed with adversarial perturbations, employing the English disturbance techniques previously mentioned. By using AdvGLUE, we are able to benchmark the performance of our models against a set of pre-defined adversarial challenges, providing a robust assessment of their resilience and adaptability in handling adversarial inputs. This dual approach ensures a comprehensive evaluation of model performance across different languages and types of perturbations.

\subsection{Datasets}

We conduct our experiments on five distinct datasets, each designed to address different types of spelling errors within the context of Chinese spelling correction:

\textbf{Perfect Pinyin}: This dataset focuses on spelling errors involving characters that are visually or phonetically similar to the correct ones. It uses the perfect form of pinyin as a reference, containing 23,502 training examples and 1,000 test examples.

\textbf{Abbreviation Pinyin}: Similar to the Perfect Pinyin dataset, this dataset deals with errors involving visually or phonetically similar characters but uses the abbreviated form of pinyin. It consists of 26,910 training examples and 1,000 test examples.

\textbf{Character Split}: This dataset addresses errors where characters are split into their component radicals or other sub-characters. It contains 25,306 training examples and 1,000 test examples.

\textbf{Hybrid}: This dataset combines the types of errors present in the Perfect Pinyin, Abbreviation Pinyin, and Character Split datasets, introducing a mix of visual and phonetic errors as well as character splitting errors. It consists of 24,839 training examples and 1,000 test examples.

\textbf{Hybrid-v2}: While based on the same types of perturbations as the Hybrid dataset, Hybrid-v2 includes additional perturbations such as insertion, deletion, inversion, and Unicode-based errors. Although it shares the same underlying perturbation methods, the data itself is different from that in the Hybrid dataset. This dataset includes 26,766 training examples and 1,000 test examples.

Each dataset includes a balanced distribution of positive and negative examples, as indicated in Table \ref{tab:7}, to ensure a robust evaluation of our methods.

\begin{table*}[htp!]
\centering
\renewcommand\arraystretch{1.0}
\resizebox{0.9\textwidth}{!}{%
\begin{tabular}{lcccccccc|cc}
\toprule
\multicolumn{1}{c|}{\multirow{3}{*}{\textbf{Model}}} & \multicolumn{2}{c|}{\textbf{TNEWS}}                 & \multicolumn{2}{c|}{\textbf{AFQMC}}                 & \multicolumn{2}{c|}{\textbf{CMNLI}}                 & \multicolumn{2}{c|}{\textbf{IFLYTEK}} & \multicolumn{2}{c}{\textbf{COLD}}             \\ \cmidrule{2-11} 
\multicolumn{1}{c|}{}                                & Clean & \multicolumn{1}{c|}{Adv}                    & Clean & \multicolumn{1}{c|}{Adv}                    & Clean & \multicolumn{1}{c|}{Adv}                    & Clean     & Adv                       & Clean                & Adv                    \\
\multicolumn{1}{c|}{}                                & (Acc) & \multicolumn{1}{c|}{(Acc)}                  & (Acc) & \multicolumn{1}{c|}{(Acc)}                  & (Acc) & \multicolumn{1}{c|}{(Acc)}                  & (Acc)     & (Acc)                     & (Acc)                & (Acc)                  \\ \midrule
\multicolumn{9}{l|}{\textit{Pre-training}}                                                                                                                                                                                                                     &                      &                        \\ \midrule
\multicolumn{1}{l|}{BERT}                            & 66.6  & \multicolumn{1}{c|}{65.4}                   & 75.1  & \multicolumn{1}{c|}{72.4}                   & 80.8  & \multicolumn{1}{c|}{77.5}                   & 58.4      & 56.2                      & 93.1                 & 80.5                   \\
\multicolumn{1}{l|}{ChineseBERT}                     & 67.0  & \multicolumn{1}{c|}{65.6}  & 73.7  & \multicolumn{1}{c|}{70.5}          & 79.9  & \multicolumn{1}{c|}{77.3}          & 57.2      & 54.9             & 93.1                 & 80.3          \\
\multicolumn{1}{l|}{MacBERT}                         & 67.6  & \multicolumn{1}{c|}{65.9}          & 74.7  & \multicolumn{1}{c|}{71.3}          & 81.0  & \multicolumn{1}{c|}{78.1}          & 58.8      & 57.5             & 93.3                 & 80.9          \\
\multicolumn{1}{l|}{RoCBERT}                         & 66.7  & \multicolumn{1}{c|}{63.7}          & 69.0  & \multicolumn{1}{c|}{69.0}          & 79.1  & \multicolumn{1}{c|}{75.7}          & 43.5      & 39.6            & 90.5                 & 79.4          \\ \midrule
\multicolumn{9}{l|}{\textit{Adversarial Tuning}}                                                                                                                                                                                                               & \multicolumn{1}{l}{} & \multicolumn{1}{l}{}   \\ \midrule
\multicolumn{1}{l|}{FreeLB}                          & 67.1  & \multicolumn{1}{c|}{65.5}          & 74.2  & \multicolumn{1}{c|}{70.9}          & 80.1  & \multicolumn{1}{c|}{77.4}          & 59.3      & 57.6             & 93.1                 & 80.5          \\
\multicolumn{1}{l|}{SMART}                           & 66.6  & \multicolumn{1}{c|}{64.7}          & 73.1  & \multicolumn{1}{c|}{70.9}          & 79.4  & \multicolumn{1}{c|}{76.3}          & 58.3      & 55.5             & 93.1                 & 80.5          \\
\multicolumn{1}{l|}{R3F}                             & 67.1  & \multicolumn{1}{c|}{65.5}          & 74.1  & \multicolumn{1}{c|}{71.0}          & 80.1  & \multicolumn{1}{c|}{77.5}          & 58.7      & 56.5             & 93.1                 & 80.5          \\
\multicolumn{1}{l|}{CreAT}                           & 66.8  & \multicolumn{1}{c|}{65.4}          & 73.4  & \multicolumn{1}{c|}{70.5}          & 79.0  & \multicolumn{1}{c|}{76.0}          & 58.9      & 57.2            & 93.1                 & 80.5          \\
\multicolumn{1}{l|}{LIMIT}                           & 66.6  & \multicolumn{1}{c|}{\textbf{66.0}} & 75.1  & \multicolumn{1}{c|}{\textbf{72.4}} & 80.8  & \multicolumn{1}{c|}{\textbf{79.1}} & 58.4      & \textbf{59.7}    & 93.1                 & \textbf{82.4} \\ \midrule
\multicolumn{9}{l|}{\textit{Large Language Models}}                                                                                                                                                                                                               & \multicolumn{1}{l}{} & \multicolumn{1}{l}{}   \\ \midrule
\multicolumn{1}{l|}{Llama-7B}                                 & 13.0  & \multicolumn{1}{c|}{10.7}                  & 43.5  & \multicolumn{1}{c|}{49.1}                  & 34.9  & \multicolumn{1}{c|}{34.9}                  & 47.6    & 48.7                     & 50.0     &   43.8                    \\
\multicolumn{1}{l|}{Baichuan2-7B}                             & 37.8  & \multicolumn{1}{c|}{31.2}                  & 40.7  & \multicolumn{1}{c|}{44.8}                  & 41.5  & \multicolumn{1}{c|}{40.4}                  & 44.5    & 46.1                     & 50.0     & 49.4                      \\
\multicolumn{1}{l|}{Baichuan2-13B}                            & 33.2  & \multicolumn{1}{c|}{27.9}                  & 69.0  & \multicolumn{1}{c|}{69.0}                  & 34.4  & \multicolumn{1}{c|}{33.5}                  & 44.8    & 44.0                     & 48.2     &  47.5              \\
\multicolumn{1}{l|}{GPT-3.5-turbo-175B}                       & 49.9  & \multicolumn{1}{c|}{47.9} & 69.0  & \multicolumn{1}{c|}{68.9} & 52.9  & \multicolumn{1}{c|}{51.0} & 49.8    & 37.1   & 51.9     & 50.0   \\ \bottomrule
\end{tabular}%
}
\caption{\label{tab:8}
Performance of the adversarial training baseline models and our method on the Chinese NLU dataset. The best results are labeled with \textbf{bold}.
}
\end{table*}

\begin{table}[t!]
\renewcommand\arraystretch{1.0}
\resizebox{\columnwidth}{!}{%
\begin{tabular}{lcccc}
\toprule
\multicolumn{1}{c|}{\multirow{2}{*}{\textbf{Model}}} & \multicolumn{1}{c|}{\textbf{\begin{tabular}[c]{@{}c@{}}Adv\\ SST-2\end{tabular}}} & \multicolumn{1}{c|}{\textbf{\begin{tabular}[c]{@{}c@{}}Adv\\ QQP\end{tabular}}} & \multicolumn{1}{c|}{\textbf{\begin{tabular}[c]{@{}c@{}}Adv\\ MNLI-m\end{tabular}}} & \textbf{\begin{tabular}[c]{@{}c@{}}Adv\\ RTE\end{tabular}} \\
\multicolumn{1}{c|}{}                                & \multicolumn{1}{c|}{(Acc)}                                                        & \multicolumn{1}{c|}{(Acc)}                                                      & \multicolumn{1}{c|}{(Acc)}                                                         & (Acc)                                                      \\ \midrule
\multicolumn{5}{l}{\textit{Adversarial Training \cite{wu2023toward}}}                                                                                                                                                                                                                                                                                                       \\ \midrule
\multicolumn{1}{l|}{BERT}                            & \multicolumn{1}{c|}{32.3}                                                         & \multicolumn{1}{c|}{50.8}                                                       & \multicolumn{1}{c|}{32.6}                                                          & 37.0                                                       \\
\multicolumn{1}{l|}{FreeLB}                          & \multicolumn{1}{c|}{31.6}                                                         & \multicolumn{1}{c|}{51.0}                                                       & \multicolumn{1}{c|}{33.5}                                                          & 42.0                                                       \\
\multicolumn{1}{l|}{R3F}                             & \multicolumn{1}{c|}{38.5}                                                         & \multicolumn{1}{c|}{40.6}                                                       & \multicolumn{1}{c|}{35.8}                                                          & 50.1                                                       \\
\multicolumn{1}{l|}{CreAT}                           & \multicolumn{1}{c|}{35.3}                                                         & \multicolumn{1}{c|}{51.5}                                                       & \multicolumn{1}{c|}{36.0}                                                          & 45.2                                                       \\
\multicolumn{1}{l|}{Match-Tuning}                    & \multicolumn{1}{c|}{51.4}                                                         & \multicolumn{1}{c|}{41.5}                                                       & \multicolumn{1}{c|}{35.5}                                                          & 47.5                                                       \\
\multicolumn{1}{l|}{LIMIT}                           & \multicolumn{1}{c|}{\textbf{66.2}}                                                & \multicolumn{1}{c|}{\textbf{78.8}}                                                       & \multicolumn{1}{c|}{\textbf{69.4}}                                                 & \textbf{84.0}                                              \\ \midrule
\multicolumn{5}{l}{\textit{Large Language Models \cite{wang2023robustness}}}                                                                                                                                                                                                                                                                                                \\ \midrule
\multicolumn{1}{l|}{BART-L (407M)}                   & \multicolumn{1}{c|}{43.9}                                                         & \multicolumn{1}{c|}{37.2}                                                       & \multicolumn{1}{c|}{41.3}                                                          & 43.2                                                       \\
\multicolumn{1}{l|}{GPT-J (6B)}                      & \multicolumn{1}{c|}{51.3}                                                         & \multicolumn{1}{c|}{41.0}                                                       & \multicolumn{1}{c|}{26.4}                                                          & 43.2                                                       \\
\multicolumn{1}{l|}{FLAN-T5-J (11B)}                 & \multicolumn{1}{c|}{59.5}                                                         & \multicolumn{1}{c|}{41.0}                                                       & \multicolumn{1}{c|}{51.2}                                                          & 43.2                                                       \\
\multicolumn{1}{l|}{GPT-NEOX (20B)}                  & \multicolumn{1}{c|}{47.3}                                                         & \multicolumn{1}{c|}{43.6}                                                       & \multicolumn{1}{c|}{40.5}                                                          & 51.9                                                       \\
\multicolumn{1}{l|}{OPT (66B)}                       & \multicolumn{1}{c|}{52.4}                                                         & \multicolumn{1}{c|}{46.1}                                                       & \multicolumn{1}{c|}{39.7}                                                          & 42.0                                                       \\
\multicolumn{1}{l|}{BLOOM (176B)}                    & \multicolumn{1}{c|}{51.3}                                                         & \multicolumn{1}{c|}{41.0}                                                       & \multicolumn{1}{c|}{26.4}                                                          & 43.2                                                       \\
\multicolumn{1}{l|}{GPT-3.5-turbo (175B)}            & \multicolumn{1}{c|}{60.1}                                                         & \multicolumn{1}{c|}{72.0}                                              & \multicolumn{1}{c|}{67.8}                                                          & 75.3                                                       \\ \bottomrule
\end{tabular}%
}
\caption{Adversarial robustness results on the AdvGLUE benchmark. We report the Accuracy on adversarial examples. The best performing scores are in bold. \label{tab:9}}
\end{table}

\begin{table*}[t!]
\resizebox{\textwidth}{!}{%
\renewcommand\arraystretch{1.0}
\begin{tabular}{c|l|l|lll|lll}
\toprule
\multirow{2}{*}{\textbf{Model/Dataset}}  & \textbf{IFLYTEK}             & \textbf{CMNLI}               & \multicolumn{3}{c|}{\textbf{ADGEN}}                                                        & \multicolumn{3}{c}{\textbf{CSL}}                                                           \\
                                         & (Acc)                        & (Acc)                        & (Rouge-1)                    & (Rouge-2)                    & (Rouge-L)                    & (Rouge-1)                    & (Rouge-2)                    & (Rouge-L)                    \\ \midrule
\multicolumn{1}{l|}{Vanilla Fine-tuning} & 56.2                         & 77.5                         & 40.6                         & 15.6                         & 23.9                         & 52.8                         & 38.2                         & 49.9                         \\
\multicolumn{1}{r|}{+SC}                 & 59.3                         & 78.8                         & 41.5                         & 16.4                         & 24.5                         & 58.0                         & 45.1                         & 55.4                         \\
\multicolumn{1}{r|}{+DI}                 & $\textbf{59.7}_{\uparrow 3.7}$ & $\textbf{79.1}_{\uparrow 1.6}$ & $\textbf{41.6}_{\uparrow 1.0}$ & $\textbf{16.7}_{\uparrow 1.1}$ & $\textbf{24.9}_{\uparrow 1.0}$ & $\textbf{58.9}_{\uparrow 6.1}$ & $\textbf{45.3}_{\uparrow 7.1}$ & $\textbf{55.5}_{\uparrow 5.6}$ \\ \bottomrule
\end{tabular}%
}
\caption{
Ablation results in different components of LIMIT. The best-performing scores are in bold. \label{tab:10}
}
\end{table*}

\subsection{Experimental Metrics}

\subsubsection{Unnatural Text Correction Metrics}

\paragraph{Precision}
Precision measures the proportion of true positive corrections among all items identified as positive. The formula is:
\[
\text{Precision} = \frac{\text{True Positives}}{\text{True Positives} + \text{False Positives}}
\]
where:
\begin{itemize}
    \item \text{True Positives (TP)}: Corrections correctly identified as positive.
    \item \text{False Positives (FP)}: Corrections incorrectly identified as positive.
\end{itemize}

\paragraph{F1 Score}
The F1 score is the harmonic mean of precision and recall, providing a single metric that balances both aspects. It is calculated as:
\[
\text{F1 Score} = 2 \times \frac{\text{Precision} \times \text{Recall}}{\text{Precision} + \text{Recall}}
\]
where:
\begin{itemize}
    \item \text{Recall}: The proportion of true positive corrections among all actual positives. It is calculated as:
    \[
    \text{Recall} = \frac{\text{True Positives}}{\text{True Positives} + \text{False Negatives}}
    \]
    \item \text{False Negatives (FN)}: Actual positives that are incorrectly identified as negative.
\end{itemize}

\subsubsection{Natural Language Understanding Metrics}

\paragraph{Accuracy}
Accuracy measures the proportion of correctly predicted items out of all predictions made. It is calculated as:
\[
\text{Accuracy} = \frac{\text{Number of Correct Predictions}}{\text{Total Number of Predictions}}
\]
where:
\begin{itemize}
    \item \text{Number of Correct Predictions}: The count of correctly predicted samples.
    \item \text{Total Number of Predictions}: The total number of predictions made.
\end{itemize}

\subsubsection{Natural Language Generation Metrics}

\paragraph{Rouge-1}
Rouge-1 measures the overlap of unigrams (single words) between the generated text and the reference text. It focuses on word-level matching. The formula is:
\[
\text{Rouge-1} = \frac{\text{Number of Overlapping Unigrams}}{\text{Number of Unigrams in Reference}}
\]
where:
\begin{itemize}
    \item \text{Number of Overlapping Unigrams}: The number of unigrams that appear in both the generated text and the reference text.
    \item \text{Number of Unigrams in Reference}: The total number of unigrams in the reference text.
\end{itemize}

\paragraph{Rouge-2}
Rouge-2 evaluates the overlap of bigrams (pairs of consecutive words) between the generated text and the reference text. The formula is:
\[
\text{Rouge-2} = \frac{\text{Number of Overlapping Bigrams}}{\text{Number of Bigrams in Reference}}
\]
where:
\begin{itemize}
    \item \text{Number of Overlapping Bigrams}: The number of bigrams that appear in both the generated text and the reference text.
    \item \text{Number of Bigrams in Reference}: The total number of bigrams in the reference text.
\end{itemize}

\paragraph{Rouge-L}
Rouge-L measures the longest common subsequence (LCS) between the generated text and the reference text. It evaluates the fluency and coverage of the generated text. The formula is:
\[
\text{Rouge-L} = \frac{\text{Length of LCS}}{\text{Length of Reference}}
\]
where:
\begin{itemize}
    \item \text{Length of LCS}: The length of the longest common subsequence between the generated text and the reference text.
    \item \text{Length of Reference}: The length of the reference text.
\end{itemize}

\subsection{Results on Chinese NLU Datasets}

Table \ref{tab:8} displays the experimental results across five Chinese Natural Language Understanding (NLU) datasets. The comparison includes not only adversarial training and large language models but also three Chinese pre-trained models: ChineseBERT, MacBERT, and RoCBERT. We specifically examine the performance variations of these models before and after the introduction of adversarial perturbations. 

This evaluation allows us to assess the robustness of each model under adversarial conditions, providing insights into their relative strengths and weaknesses. By analyzing how performance changes in response to adversarial challenges, we gain a deeper understanding of each model’s effectiveness and resilience in Chinese NLU tasks, thus informing their practical applicability and reliability.

MacBERT employs a pre-training and fine-tuning paradigm similar to that of BERT. The well-designed pre-training for Chinese allows it to outperform BERT on both clean and adversarial datasets from TNEWS, CMNLI, and IFLYTEK. Regarding contrastive learning, RoCBERT performs relatively well in the clean TNEWS and CMNLI tasks. However, its performance on adversarial TNEWS, AFQMC, CMNLI, and IFLYTEK drops by 1.7\%, 3.4\%, 1.8\%, and 16.6\%, respectively. The experimental results reflect that it over-focuses on robust representations while suffering performance loss.
The experimental demonstrates show that the adversarial robustness of our proposed LIMIT achieves consistent improvement on the NLU task. On the perturbed TNEWS, AFQMC, CMNLI, IFLYTEK, and COLD datasets, LIMIT obtains an improvement of 0.6\%, 1.9\%, 3.1\%, 2.5\%, and 1.9\%, respectively. We find that all the adversarial training methods suffer a loss of performance in Chinese tasks. The trade-off between performance and robustness is consistent with previous findings. For LIMIT, however, it is only responsible for removing adversarial perturbations from the input text. This preserves the performance of the language model to some extent.
For example, the FreeLB and R3F outperform vanilla BERT on clean TNEWS and IFLYTEK datasets, while the SMART and CreAT sacrifice prediction performance on all tasks. On the relatively easier classification adversarial datasets TNEWS and IFLYTEK, most of the methods provide performance gains. However, for the inference tasks AFQMC and CMNLI, both lead to performance loss when trading off performance and robustness. 

\subsection{Results on English NLU Datasets}

Table \ref{tab:9} shows the experimental results on four English NLU adversarial datasets (AdvGLUE). Experimental results illustrate that LIMIT outperforms state-of-the-art methods and achieves the best performance on four randomly selected datasets. The appendix provides additional details on various types and parameter scales of large language models.
For the pre-training and fine-tuning methods, Match-Tuning with BERT-large achieves competitive results by adding regularization. For the large language models, ChatGPT exhibits better performance than the specifically designed model, achieving accuracy scores of 60.1\%, 72.0\%, 67.8\%, and 65.5\% on the four datasets, respectively. However, models with the same parameter sizes show considerable variation in performance, with an average accuracy of only 42.2\% for BLOOM. For adversarial training methods, FreeLB, R3F, and CreAT perform poorly, which has validated their struggles to cope with multitype errors and perturbations.

\subsection{Ablation Study}

Table \ref{tab:10} shows the ablation studies of the different components of LIMIT on the Chinese and English adversarial datasets. It indicates that the components of self-correct adversarial training (SC) and decoding intervention (DI) both play key roles in enhancing adversarial robustness. Specifically, with the addition of SC, the average accuracy of the NLU dataset is improved by an average of 2.2\%, and the Rouge-L of the NLG dataset is improved by an average of 3.1\%. Similarly, with the addition of DI, the average accuracy of the NLU dataset is improved by an average of 0.4\%, and an average of 0.3\% improves the Rouge-L of the NLG dataset.

\bibliography{aaai25}

\end{document}